\documentclass[sigconf]{acmart}

\AtBeginDocument{%
  \providecommand\BibTeX{{%
    \normalfont B\kern-0.5em{\scshape i\kern-0.25em b}\kern-0.8em\TeX}}}

\setcopyright{acmcopyright}

\copyrightyear{2022}
\acmYear{2022}
\setcopyright{acmlicensed}\acmConference[MM '22]{Proceedings of the 30th ACM International Conference on Multimedia}{October 10--14, 2022}{Lisboa, Portugal}
\acmBooktitle{Proceedings of the 30th ACM International Conference on Multimedia (MM '22), October 10--14, 2022, Lisboa, Portugal}
\acmPrice{15.00}
\acmDOI{10.1145/3503161.3548419}
\acmISBN{978-1-4503-9203-7/22/10}

\usepackage{caption}
\usepackage{balance}

\begin{document}

\title{xCloth: Extracting Template-free Textured 3D Clothes from a Monocular Image}


\author{Astitva Srivastava}
\email{astitva.srivastava@research.iiit.ac.in}
\affiliation{%
  \institution{Center for Visual Information Technology, IIIT Hyderabad}
  \streetaddress{Gachibowli}
  \city{Hyderabad}
  \state{Telangana}
  \country{India}
  \postcode{500032}
}

\author{Chandradeep Pokhariya}
\email{chandradeep.pokhariya@research.iiit.ac.in}
\affiliation{%
  \institution{Center for Visual Information Technology, IIIT Hyderabad}
  \streetaddress{Gachibowli}
  \city{Hyderabad}
  \state{Telangana}
  \country{India}
  \postcode{500032}
}

\author{Sai Sagar Jinka}
\email{jinka.sagar@research.iiit.ac.in}
\affiliation{%
  \institution{Center for Visual Information Technology, IIIT Hyderabad}
  \streetaddress{Gachibowli}
  \city{Hyderabad}
  \state{Telangana}
  \country{India}
  \postcode{500032}
}

\author{Avinash Sharma}
\email{asharma@iiit.ac.in}
\affiliation{%
  \institution{Center for Visual Information Technology, IIIT Hyderabad}
  \streetaddress{Gachibowli}
  \city{Hyderabad}
  \state{Telangana}
  \country{India}
  \postcode{500032}
}






\renewcommand{\shortauthors}{Astitva Srivastava, Chandradeep Pokhariya, Sai Sagar Jinka, \& Avinash Sharma}


\begin{abstract}
Existing approaches for 3D garment reconstruction either assume a predefined template for the garment geometry (restricting them to fixed clothing styles) or yield vertex colored meshes (lacking high-frequency textural details). Our novel framework co-learns geometric and semantic information of garment surface from the input monocular image for template-free textured 3D garment digitization. More specifically, we propose to extend PeeledHuman representation to predict the pixel-aligned, layered depth and semantic maps to extract 3D garments. The layered representation is further exploited to UV parametrize the arbitrary surface of the extracted garment without any human intervention to form a UV atlas. The texture is then imparted on the UV atlas in a hybrid fashion by first projecting pixels from the input image to UV space for the visible region, followed by inpainting the occluded regions. Thus, we are able to digitize arbitrarily loose clothing styles while retaining high-frequency textural details from a monocular image. We achieve high-fidelity 3D garment reconstruction results on three publicly available datasets and generalization on internet images.


\end{abstract}





\begin{teaserfigure}
\begin{center}
  \includegraphics[width=\textwidth]{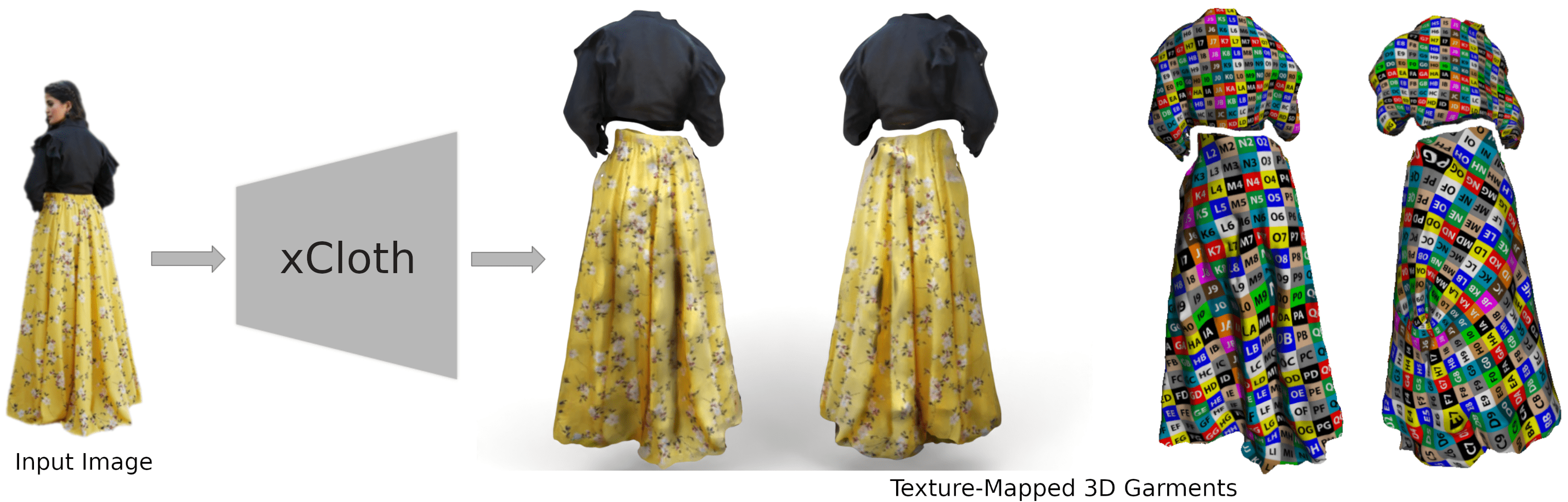}
  \caption{Proposed xCloth framework extracts high-fidelity template-free textured 3D garments from a monocular image.}
  \Description{The teaser figure}
  \label{fig:teaser}
\end{center}
\end{teaserfigure}


\maketitle

\section{Introduction}
\label{sec:introduction}
The high-fidelity digitization of 3D garment(s) from a 2D image(s) is essential in achieving photorealism in a wide range of applications in computer vision, e.g., 3D virtual try-on, human digitization for AR/VR, realistic virtual avatar animation, etc. The goal of the 3D digitization is to recover 3D surface geometry as well as high-frequency textural details of the garments with arbitrary styles/designs.
Textured 3D digitization of garments remains notoriously tricky, as designing clothing is a work of art where a lot of creativity is involved at the designer's end. The lack of standardization in designing vocabulary (e.g., cloth panels) further makes the task challenging. 
Traditionally, expensive multi-view capture/scan setups were used for digitizing garments, but they can’t scale up in fast-fashion scenarios ~\cite{bhardwaj2010fast,gazzola2020trends}, owing to their cost, scan latency and volume of efforts.

The majority of the existing learning-based garment digitization methods~\cite{yang2018physics,mir20pix2surf,majithia2022robust} rely on the availability of predefined 3D template mesh for a specific clothing style, generally taken from popular parametric body models (e.g., SMPL ~\cite{Loper2015SMPLAS}) or designed by an artist ~\cite{BerkeleyGarment}. Specifically, one class of methods (e.g.,~\cite{mir20pix2surf, majithia2022robust}) propose to transfer the texture from the RGB image to the UV map of the predefined template while retaining the fixed template geometry. The other class of methods (e.g.,~\cite{bhatnagar2019mgn, Jiang2020BCNetLB, Zhu2020DeepFA}) propose to deform the predefined template using cues from input RGB image in a learnable fashion but do not attempt to recover texture. MGN ~\cite{bhatnagar2019mgn} proposed a hybrid approach to achieve the best of these two by learning to locally deform the template mesh using SMPL+D for recovering geometry and transfer texture from multiple images to SMPL UV atlas using \cite{alldieck2019learning} which is fixed for any garment style.
However, these methods restrict the clothing styles to a set of predefined templates, which are very simplistic in nature and cannot model arbitrary clothing styles with high-frequency geometrical details (e.g., folds in long skirts). A recent attempt in~\cite{Zhu2020DeepFA} propose to modify the SMPL template by adaptable template using handle-based deformation and register it to non-parametric mesh recovered from OccNet~\cite{mescheder2019occupancy} to accommodate loose clothing. Another very recent attempt \cite{https://doi.org/10.48550/arxiv.2203.15007}   further extends this idea to incorporate learnable segmentation \&  garment-boundary field to improve the reconstruction quality. However, these methods do not attempt to recover textured or even vertex-colored meshes. A similar scenario exists in the related domain of clothed human body reconstruction, where all existing methods either recover 3D mesh surface sans texture or achieve a fix (SMPL mesh) texture map based reconstruction, which imposes tight clothing limitations.

\begin{figure}[b]
    \centering
  \includegraphics[width=0.5\textwidth]{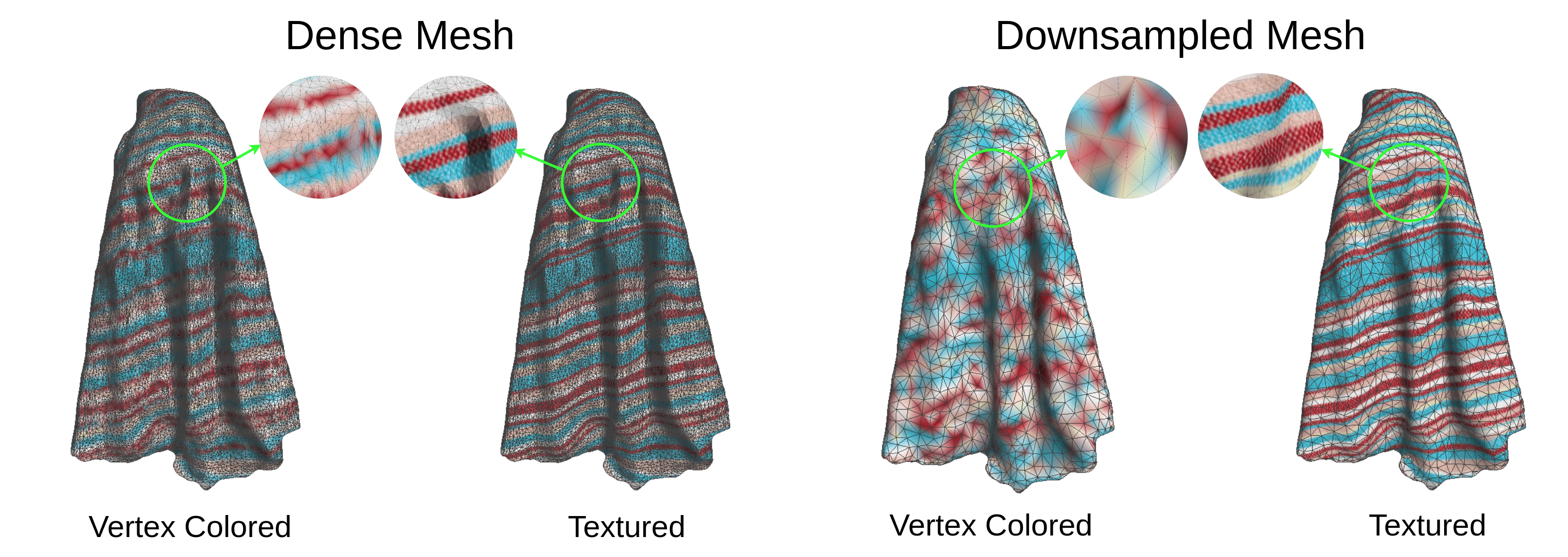}
  \caption{Advantage of textured over vertex-colored mesh.}
  \Description{The figure depicts the difference between vertex colored and textured mesh.}
  \label{fig:textvsvcolor}
\end{figure}

A textured mesh not only retains high-quality appearance details that are crucial for a 3D virtual-try-on kind of application, but it can also be exploited for extended applications such as texture map super-resolution for enhanced detailing, appearance manipulation by texture swapping, etc. %
In regard to computational efficiency, texture representation is also memory efficient as high-frequency appearance details can be retained by rendering low-quality meshes (with less number of faces), as shown in Fig.~\ref{fig:textvsvcolor}. 
Existing template-based garment digitization methods assumes a predefined UV parametrized template for texture mapping. 
%
Various parameterization techniques are available (e.g., LSCM \cite{levy2002least}) that can be used for template-free garment reconstruction. However, they need to find optimal seams (for partitioning the mesh), which often require either manual intervention or do not provide any control over the placements of seams.
%
%
%
%

In this paper, we propose a novel framework for template-free, textured 3D digitization of arbitrary garment styles from a monocular image. The proposed framework consists of three modules; where first, we predict the 3D geometry of the garment in the form of a sparse, non-parametric peeled representation (to handle self-occlusions) along with semantic information and surface normals in the form of semantic and normal peelmaps.  
The semantic labels provide supervision for extracting the cloths separately and also help in dealing with complex garment geometry and pose, while the normal maps help in recovering the high-frequency surface details. This yields a dense point-cloud representing the garment. Subsequently, we refine this point cloud to recover a mesh representation of the garment. Finally, we exploited the peeled representation in a novel way to automatically UV parametrize the extracted 3D garment mesh. More specifically, the peeled representation provides natural view-specific partitions of the 3D garment mesh; we propose to automatically infer seams by exploiting these partitions for UV parametrization.
%
%
We recover the associated texture map by appropriately projecting the RGB peelmaps to the corresponding partitions.
However, in many cases where the garment has high-frequency textural details, generative methods tend to yield blurred predictions and hence, as a remedy, we propose to use an existing planar structure-based inpainting method for the texture maps. We thoroughly evaluated the proposed framework, reported qualitative and quantitative results on the publicly available datasets as well as internet images and reported superior performance as compared to existing SOTA methods. We intend to release the proposed model in the public domain. 

Please refer to the supplementary material for a detailed literature survey.
\section{Proposed Framework}
\label{sec:method}
The proposed xCloth framework is divided into three key modules. First, the input image is passed to the "3D Reconstruction" module, which extracts the 3D garment in the form of a point cloud. This point cloud contains missing regions, especially along the tangential boundaries, owing to the inherent limitation of peeled representation. The noisy point cloud is passed to the "Geometry Refinement" module, which deals with the missing regions and extracts a dense mesh out of it. Finally, the mesh is passed to the "Peeled Texture Mapping \& In-painting" module, where the mesh is automatically (UV-) parametrized, followed by texture in-painting for the occluded regions producing a high-fidelity textured mapped 3D garment. We discuss each of these modules in detail below.
%
%
\begin{figure*}
    \centering
  \includegraphics[width=0.70\textwidth]{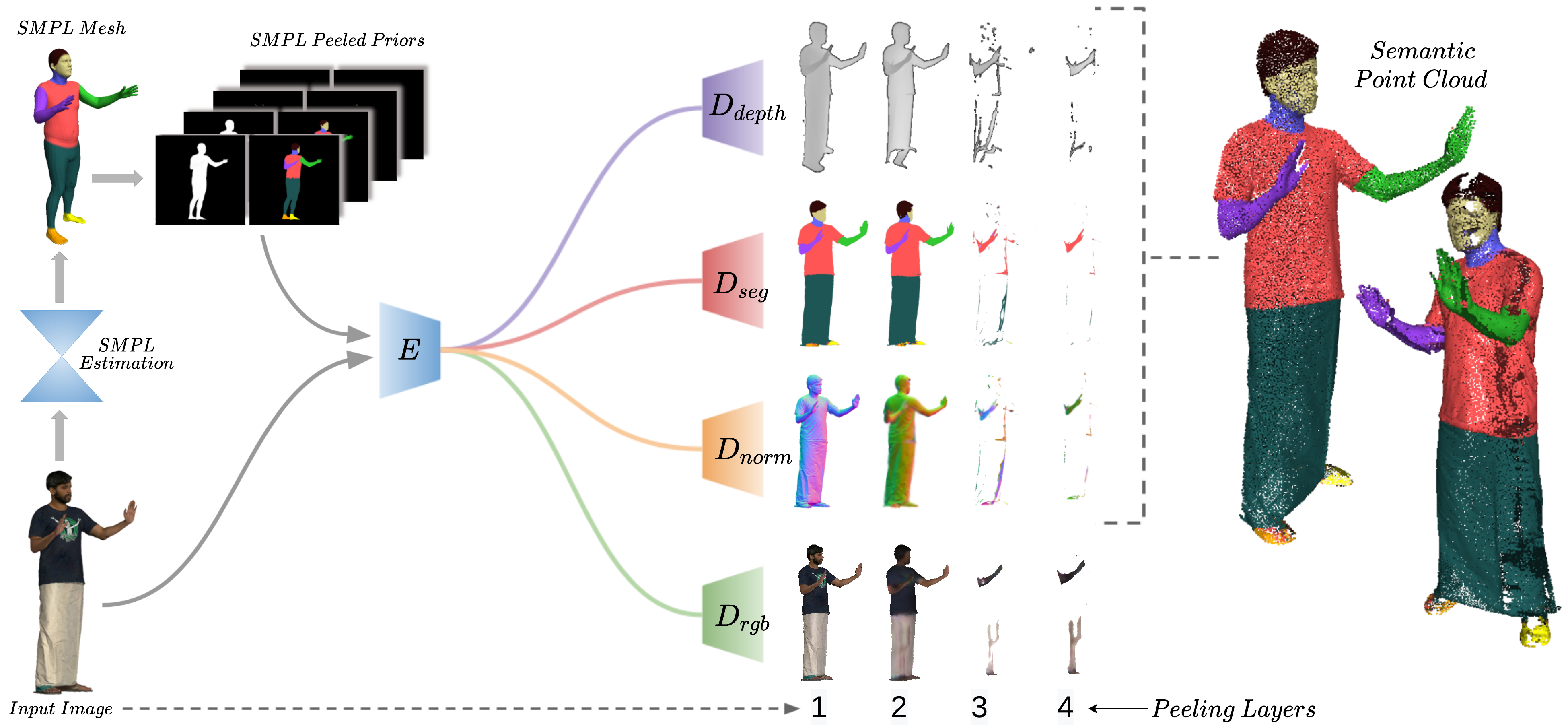}
  \caption{Outline of the proposed 3D Reconstruction Module. SMPL mesh is estimated from the input image which is then segmented and peeled to generate SMPL peeled priors. These priors are passed to the shared encoder $\mathbf{\textit{E}}$. The output of encoder is passed to the decoders \textit{D\textsubscript{depth}}, \textit{D\textsubscript{seg}}, \textit{D\textsubscript{norm}} \& \textit{D\textsubscript{rgb}}, which produce depth, segmentation, normal and RGB peelmaps. The semantic point cloud is generated from depth, segmentation and normal peelmaps.}
  \Description{The figure depicts proposed 3D Reconstruction Module.}
  \label{fig:module_1}
\end{figure*}
\subsection{3D Reconstruction}
\label{sec:reconstruction}
The first task is to recover the 3D geometry of the garment from a monocular image. To represent the 3D geometry, we use PeeledHuman \cite{jinka2020peeledhuman} representation which is a non-parametric, multi-layered encoding of 3D shapes, stored in the form of sparse images called peelmaps. To generate ground truth peelmaps, the mesh is placed in a virtual scene and a set of rays are emanated from the camera center through each pixel towards the mesh. The first set of ray-intersections with the mesh are recorded as the first layer depth peelmap and RGB peelmap, capturing visible surface information nearest to the camera. Subsequently, the rays are extended beyond the first intersection point (piercing through the intersecting surface) to hit the surface behind it. The corresponding depth and RGB values are recorded in the next layer peelmaps. We use total four layers of peeled representation for the 3D reconstruction module. 
We adopt the architecture proposed in \cite{sharp}, which is comprised of a shared encoder and predicts only RGB and Depth peelmaps. We extend the proposed architecture as follows. The shared encoder encodes the input image along with peeled SMPL (depth and body part segmentation) priors. This common encoding is fed to four decoder branches, namely, $D_{depth}$, $D_{seg}$, $D_{norm}$ \& $D_{rgb}$, which predicts depth peelmaps $\widehat{\mathcal{P}}_{depth}^i$, segmentation peelmaps $\widehat{\mathcal{P}}_{seg}^i$ , normal peelmaps $\widehat{\mathcal{P}}_{norm}^i$ \& RGB peelmaps $\widehat{\mathcal{P}}_{rgb}^i$, respectively.

We compute L1 loss on $\widehat{\mathcal{P}}_{depth}^i$ \& $\widehat{\mathcal{P}}_{rgb}^i$, while L2 loss on $\widehat{\mathcal{P}}_{norm}^i$ w.r.t corresponding ground truth peelmaps $\mathcal{P}$, represented by following equations:
\begin{equation}
            L_{depth} = \sum_{i=1}^{4} \Big\lVert {\widehat{ \mathcal{P}}_{depth}^i - \mathcal{P}_{depth}^i}\Big\lVert
\end{equation}
\begin{equation}
            L_{rgb} = \sum_{i=1}^{4} \Big\lVert {\widehat{ \mathcal{P}}_{rgb}^i - \mathcal{P}_{rgb}^i}\Big\lVert
\end{equation}
\begin{equation}
            L_{norm} = \sum_{i=1}^{4} {( {\widehat{ \mathcal{P}}_{norm}^i - \mathcal{P}_{norm}^i})}^2
\end{equation}

In the case of segmentation peelmaps, we minimize the multi-class cross-entropy loss over the segmentation label classes, represented by the following equation:
\begin{equation}
            L_{seg} = \sum_{c=1}^{N} { \mathcal{P}_{seg}^c \mathcal{\log}(\widehat{P}_{seg}^c)}
\end{equation}
\begin{figure}[b]
    \centering
  \includegraphics[width=0.5\textwidth]{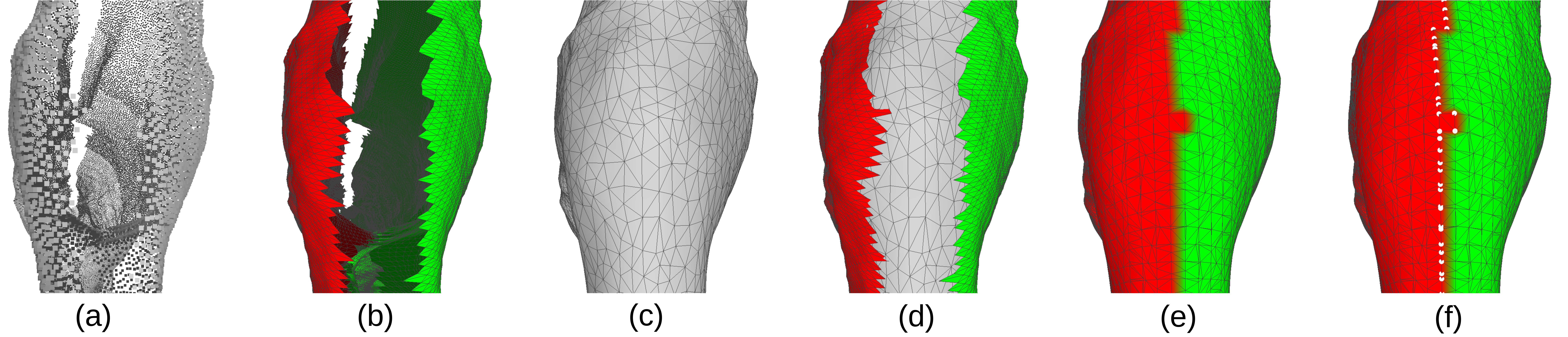}
  \caption{Geometry refinement \& automatic seam estimation.}
  \Description{The figure depicts the working of geometry refinement module and automatic to seam detection process.}
  \label{fig:geo-refine-seam}
\end{figure}
Here, $N$ refers to the total number of semantic segmentation labels, which correspond with the segmentation classes present in the CIHP\cite{gong2018instancelevel}  dataset. Please refer to supplementary material for the ground truth semantic data generation process. We train the entire network in an end-to-end fashion, minimizing the following objective function:
\begin{equation}
    L = \lambda_{depth}L_{depth} +  \lambda_{seg}L_{seg} + \lambda_{norm}L_{norm} + \lambda_{rgb}L_{rgb}
\end{equation}

The output of the 3D reconstruction module is the point cloud estimated from back-projected depth peelmaps, where each 3D point is assigned a segmentation label from the predicted semantic peelmaps, and point normals are assigned from the predicted normal peelmaps. The semantic labelling information is further used to extract each garment separately in the form of a point cloud. The predicted RGB peelmaps are later used during the texture-mapping.
\subsection{Geometry Refinement}
\label{sec:geometry-refinement}
The reconstruction module yields a dense point cloud representing the garment that has missing regions due to the inherent drawback of PeeledHuman representation, as shown in \autoref{fig:geo-refine-seam}(a). These missing regions are typically the regions that are tangential to the peeling camera rays (w.r.t. input view) and hence need to be filled to form a complete dense 3D surface of the garment. One common solution is to run the Poisson Surface Reconstruction (PSR) on top of the point cloud, which gives a watertight and hole-filled mesh. However, PSR also smoothens the geometrical details present in the back-projected depth peelmaps. Instead of naively running PSR, we first induce independent partial mesh (\autoref{fig:geo-refine-seam}(b)) for each of the depth peelmaps by exploiting the image grid structure. Additionally, for the missing region, we induce another partial mesh by sampling 3D points on the Poisson surface reconstruction of the point cloud. Finally, we merge these partial meshes to obtain a single refined mesh as in \autoref{fig:geo-refine-seam}(c). This helps in retaining the fine-grained surface details present in the predicted depth maps while filling holes \& missing regions. An additional advantage of independent meshification of each peelmaps is that we can retain the association of each vertex in the final mesh with the respective peeling layer. This association information is subsequently used for automatic seam estimation for texture mapping in the next module.

\subsection{Peeled Texture Mapping \& In-painting}
\label{sec:texture-mapping}
%



%
%
The proposed peeled texture mapping approach has three steps:

\textbf{1. Automated Seam Estimation:} Traditionally, seams are used to partition the mesh for obtaining non-overlapping UV parametrization. 
We propose a novel method to automatically estimate seams (without any manual intervention) by exploiting the partitioning provided by depth peelmaps. 
The previous module assigns each vertex of the refined mesh to a specific peeling layer, except for the vertices belonging to filled regions (\autoref{fig:geo-refine-seam}(d)). We assign these vertices to a peeling layer using nearest neighbor extrapolation. 
Thus, every vertex on the refined mesh is assigned to a peeling layer, as can be seen in \autoref{fig:geo-refine-seam}(e). Among them, all the boundary vertices define seams (\autoref{fig:geo-refine-seam}(f)) that are used to split the refined mesh into multiple partitions. The seam vertices are replicated across the adjacent partitions in order to avoid the artifacts near the seam boundary during rendering. 
Finally, each of the partition is UV parametrized separately. 

%
%
%
\textbf{2. Peeled Parametrization:}
We employ the Boundary-First Flattening (BFF) \cite{sawhney2017boundary} parametrization technique separately on each individual partition to avoid any overlap. This gives us a UV parametrization for each partition with very minimal distortion.
Instead of mapping all the vertices of the mesh to the same UV space, we now map the vertices of each partition of the mesh to the individual UV maps, thereby constructing a UV Atlas.

\textbf{3. Texture Filling \& In-painting:} 
We use the information from RGB peelmaps for filling textures in the constructed UV atlas.  Every parametrized partition has an associated RGB peelmap pixel aligned with it. In order to fill the UV map associated with this partition, we assign the RGB value for each pixel by projecting them from the UV space to the corresponding RGB peelmap. This enables filling a large part of the UV map. However, the pixels associated with the tangential regions are not visible in the corresponding RGB peelmap and remain unfilled. We propose to employ an off-the-shelf structure-based image inpainting solution \cite{huang2014image} and fill the texture for these missing pixels independently for each UV map.
\begin{table*}[h]
\begin{minipage}[b]{0.5\textwidth}
\centering
\begin{tabular}{c|ccc|ccc}
\toprule
      &       & \multicolumn{1}{c}{Topwear} &      & \multicolumn{3}{c}{Bottomwear} \\
\midrule
Dataset & P2S $\downarrow$ & IOU $\uparrow$ & NRE $\downarrow$ & P2S $\downarrow$ & IOU $\uparrow$ & NRE $\downarrow$\\ \hline
3DHumans & {0.0087} & {0.82} & {0.089}& {0.0077} & {0.83} & {0.081} \\
THUman2.0 & {0.0079} & {0.80} & {0.094} & {0.0072} &{0.74} & {0.085} \\
DW & {0.0091} & {0.91} & {0.088} & {0.0083} & {0.95} & {0.073} \\
\bottomrule
\end{tabular}%
\captionof{table}{Quantitative evaluation of our framework.}
\label{tab:quant_eval}
\end{minipage}\qquad
%
\begin{minipage}[b]{0.4\textwidth}
\centering
\begin{tabular}{c|cc|cc}
\toprule
      & \multicolumn{2}{c|}{Topwear} & \multicolumn{2}{c}{Bottomwear} \\
\midrule
Dataset & P2S $\downarrow$ & NRE $\downarrow$ & P2S $\downarrow$ & NRE $\downarrow$\\ \hline
3DHumans & {0.0091} & {0.118} & {0.0082}& {0.122}\\
THUman2.0 & {0.0088} & {0.182} & {0.0078} & {0.177}\\
DW & {0.0096} & {0.107} & {0.0089} & {0.097}\\
\hline
\end{tabular}%
\captionof{table}{Ablation without normal peelmaps.}
\label{tab:ablate_normal_loss}
\end{minipage}
\end{table*}
%
%

Additionally, we empirically observed that predicting high-quality texture for unseen areas is a tedious task for any generative model and tends to cause artifacts and over-smoothening, which also applies to the predicted RGB peelmaps. Thus, if the image has high-frequency textures, we propose to employ the Structure-Completion inpainting for the second or subsequent RGB peelmaps by directly extending the first UV map or extending a known patch either taken from the visible region of the first peel map (input image) or from the externally provided image of the underlying cloth. This results in a completely filled UV atlas.
\section{Experiments \& Results}
\subsection{Implementation Details}
We employ a multi-branch encoder-decoder network as a part of our framework, which is trained in an end-to-end fashion. The network takes the input image concatenated with SMPL peeled priors in 512 × 512 resolution. The shared encoder consists of a convolutional layer and 2 downsampling layers which have 64, 128, and 256 kernels of size 7×7, 3×3 and 3×3, respectively. This is followed by ResNet blocks which take downsampled feature maps of size 128×128×256. The decoders consist of two upsampling layers followed by a convolutional layer, having the same kernel sizes as the shared encoder. For $D_{depth}$ and $D_{norm}$, \textit{sigmoid} activation function is used while for $D_{rgb}$, \textit{tanh} activation function is used. We use the Adam optimizer with an exponentially decaying learning rate starting from $5$ × $10^{-4}$. The training takes around 18 hrs (for 20 epochs) on four Nvidia GTX 1080Ti GPUs with a batch size of 4 and hyperparameters  $\lambda_{depth}$, $\lambda_{seg}$, $\lambda_{norm}$ \& $\lambda_{rgb}$ are empirically set to $1.0$, $0.1$, $1.0$ \& $0.05$.
\begin{figure*}
    \centering
  \includegraphics[width=0.8\textwidth]{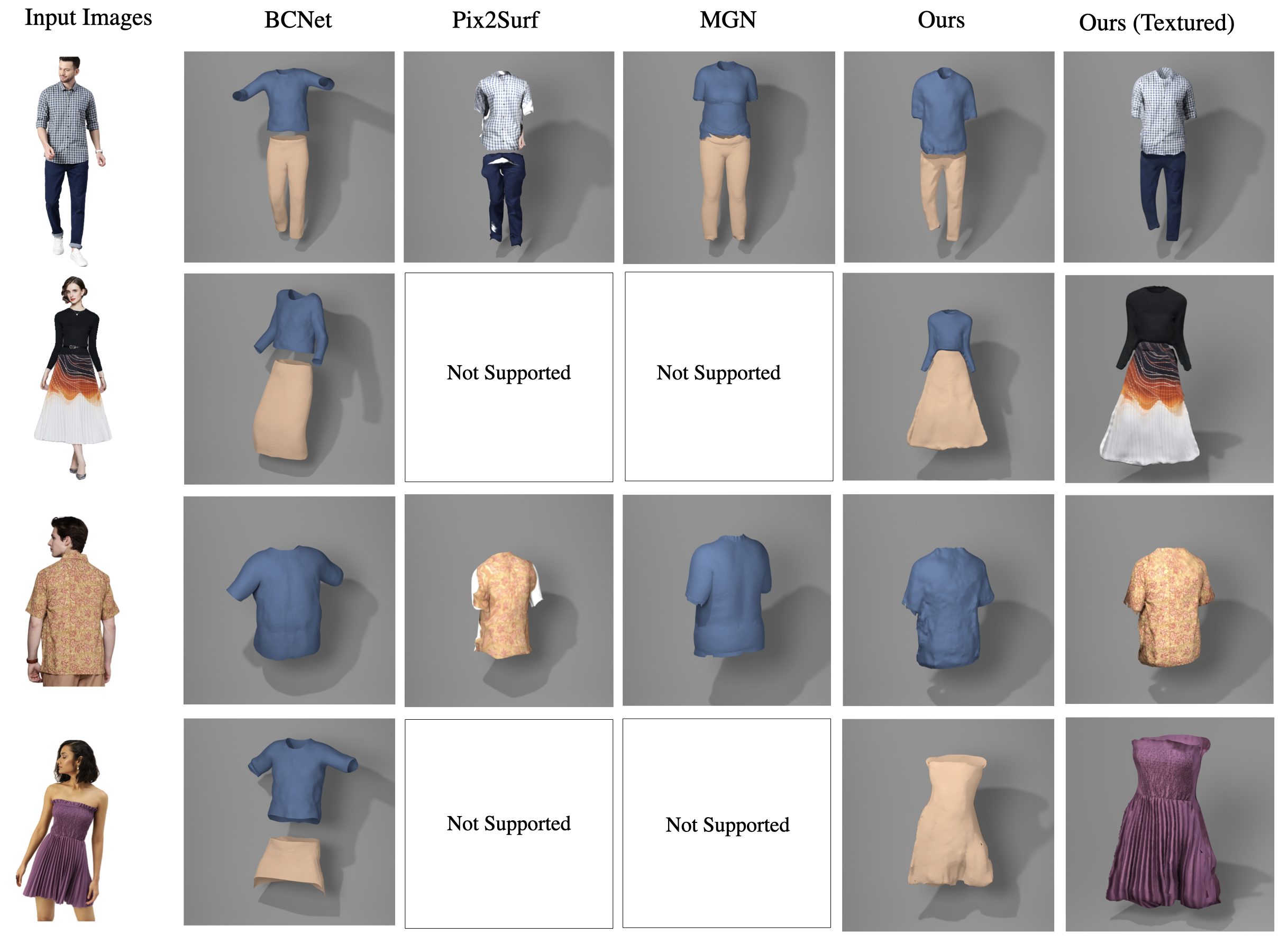}
  \caption{Qualitative comparison of xCloth (ours) with the existing SOTA methods on in-the-wild images.}
  \Description{This figures shows the comparison of our method with the all other SOTA methods. }
  \label{fig:main_comparison}
\end{figure*}
\subsection{Evaluation Metrics}
%
 \textbf{Point-to-Surface (P2S) Distance:} P2S measures the average L2 distance between each point to the surface for a given pair of point clouds and the surface. Thus, a lower value of P2S indicates higher fidelity geometric reconstruction w.r.t. ground truth surface. 
 \\
 \noindent
 \textbf{Intersection Over Union (IOU):} It is the area of overlap between the predicted and the ground truth segmentation labels divided by the area of union between them. Thus, this metric ranges from 0 to 1, and values closer to 1 are desired, indicating a higher overlap between predicted and ground truth labels.
 \\
\noindent
\textbf{Normal Reprojection Error (NRE)\cite{saito2019pifu}:} 
NRE is computed by estimating L2 error between rendered normal maps (from input view) for reconstructed and ground truth surfaces. Hence, a lower value is desired for NRE, indicating higher fidelity of reconstructed surface normals. 
\subsection{Datasets}
We use following datasets for training \& evaluation. 
\\
\noindent
 \textbf{Digital Wardrobe (DW)~\cite{bhatnagar2019mgn}:}  The dataset consist of 356 real body scans with textured meshes in arbitrary, yet simple poses and include only five categories of tight clothing styles. The dataset also includes underlying SMPL body and segmentation labels. However, out of 365 meshes, only 96 are made publicly available. We train our 3D reconstruction module on 80 out of 96 meshes and report evaluation metrics on the remaining 16 meshes.
 \\
\noindent
 \textbf{THUman2.0~\cite{yu2021function4d}:} This has around 500 real body scans with arbitrary clothing and diverse poses. We register the SMPL to the provided scans and also curate segmentation labels using our scan-segmentation pipeline. We train our 3D reconstruction module on 400 meshes and test on remaining 100 meshes.
 \\
\noindent
 \textbf{3DHumans ~\cite{3DHumansDataset,sharp}:} This recently introduced dataset contains 200 high-frequency textured scans of the South-Asian population in diverse poses with garments of arbitrary loose styles. We post-process this dataset in a similar fashion as THUman2.0. Around 170 meshes are used for training and 30 for testing.

We intend to release the curated segmentation labels for THUman2.0 and 3DHumans dataset with our scan-segmentation pipeline.
%
\begin{figure*}
    \centering
  \includegraphics[width=0.60\textwidth]{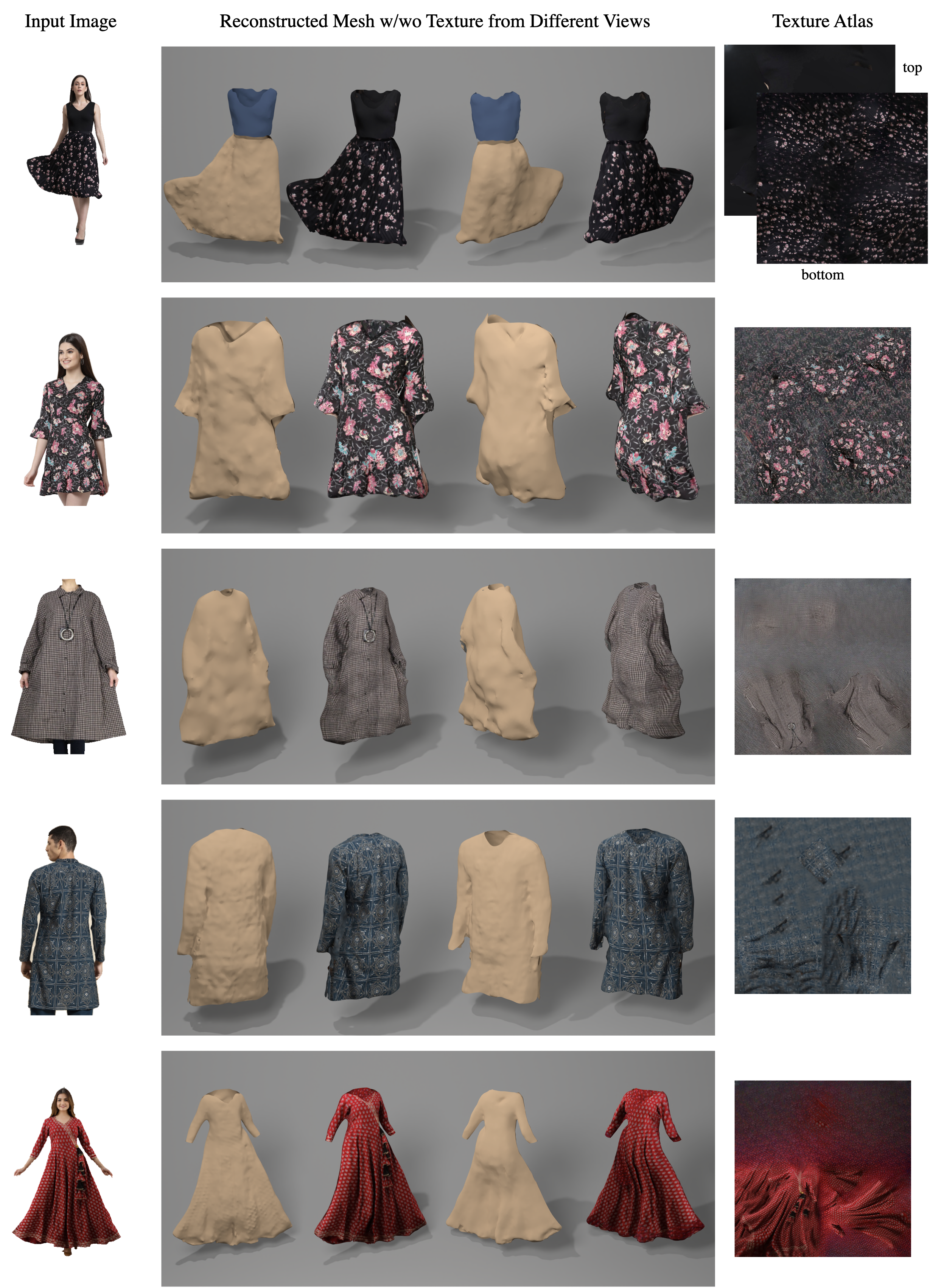}
  \caption{Results generated by xCloth for in-the-wild internet images. For each row, the input image is followed by the reconstruction in a different view (including geometry and textured mesh) and respective texture maps.}
  \Description{This figures shows the results of our method in the internet images. }
  \label{fig:ourresults_internet}
\end{figure*}
\subsection{Quantitative Evaluation}
\begin{figure}[t]
    \centering
  \includegraphics[trim={0 2cm 0 1cm},clip, width=0.5\textwidth]{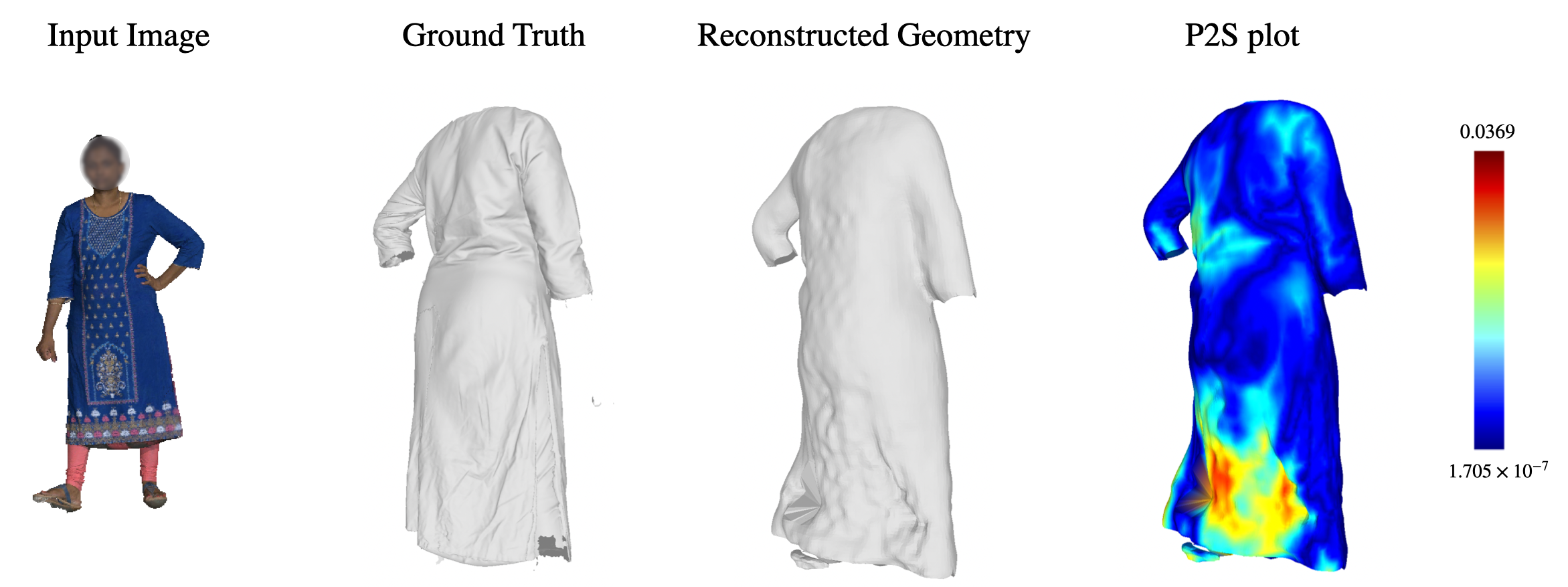} 
  \caption{Visualization of P2S distance.} 
  \Description{The figure depicts the P2S plot of our prediction.}
  \label{fig:p2splot}
\end{figure}

We evaluate xCloth framework separately for two classes of garments (topwear \& bottomwear). To evaluate the reconstruction error, we uniformly sample points on the extracted 3D garment mesh and compute P2S distance with the ground truth garment mesh. For evaluating the output of the segmentation branch, we report per-class IOU (i.e. topwear and bottomwear) for all segmentation peelmaps. Finally, to evaluate the quality of the extracted garment surface, we compute NRE separately for each class (ignoring the background). We report the quantitative results in \autoref{tab:quant_eval}. Here, we can infer that the DW dataset has lower NRE as opposed to the other two datasets. This can be attributed to the fact that it has meshes that largely lack fine-grained geometrical details in comparison to THUman2.0 and 3DHumans. On the other hand we observe that P2S distance reduces with larger training data size and thus our framework obtain higher P2S on the DW dataset as compare to other two datasets.  
We take a random test sample from the 3DHumans dataset and visualize  the P2S distance (w.r.t., corresponding ground truth mesh) on the self-occluded (back side) of the garment in \autoref{fig:p2splot}. As we can infer that reconstruction error is slightly higher in the region where clothing is loose and far from the body surface. This is due to the inherently ill-posed nature of the problem, even though the final reconstructed garment mesh looks plausible.

In regard to comparison with SOTA methods, the training code for BCNet \cite{Jiang2020BCNetLB} \& MGN \cite{bhatnagar2019mgn}) is not publicly available, and Pix2Surf \cite{mir20pix2surf} does not model geometry of the garment and hence we skip comparing with them. DeepFashion3D \cite{Zhu2020DeepFA} is a closely related work to xCloth in terms of geometry estimation as it can model arbitrary geometry up to some extent; however, the authors do not provide training/inference code. Since OccNet \cite{mescheder2019occupancy} is a major part of DeepFashion3D and the authors compare the method with OccNet in the paper, we train OccNet \cite{mescheder2019occupancy} on 3DHumans and THUman2.0 datasets and quantitatively compare it with xCloth in terms of P2S. In OccNet, neither there is any provision for semantic segmentation for each garment class, nor the final reconstruction is pixel-aligned with the input image; therefore, we skip IOU and NRE during quantitative comparison. We report the quantitative comparison with OccNet in \autoref{tab:quant_compare_occnet} where our framework significantly outperforms OccNet for both Topwear and Bottomwear. This can be attributed to the fact that OccNet relies on global features while our framework exploits pixel-aligned features. 
\begin{figure}[b]
    \centering
  \includegraphics[width=0.5\textwidth]{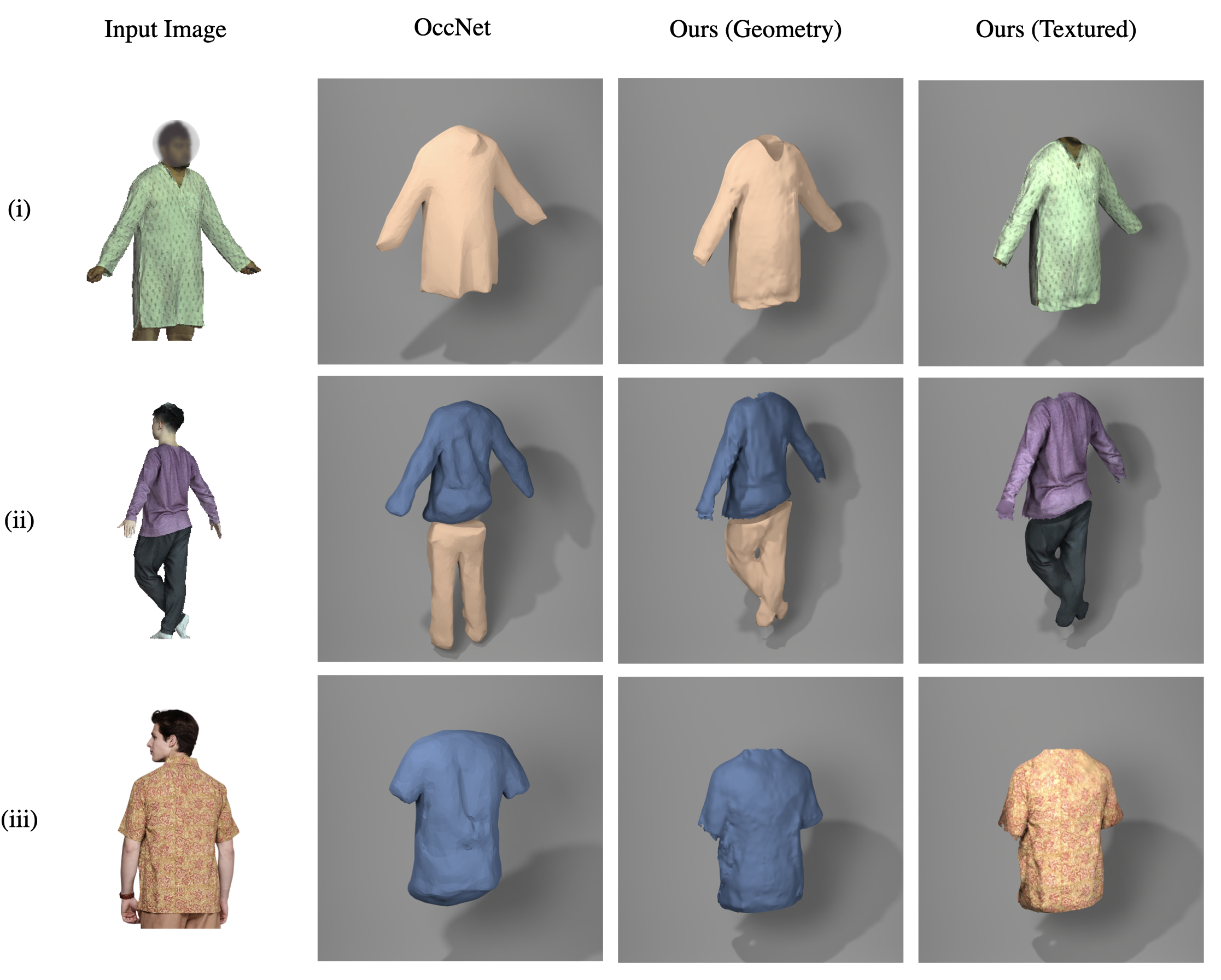}
  \caption{Qualitative comparison with OccNet \cite{mescheder2019occupancy}.}
  \Description{This figures shows the comparison of our method with the OccNet}
  \label{fig:comparison_occnet}
\end{figure}
\begin{table}[t]
%
\begin{tabular}{c|c|c|c}
\midrule
Dataset & Method & P2S $\downarrow$ (Topwear)& P2S $\downarrow$ (Bottomwear)\\ \hline
3DHumans & 
\begin{tabular}{c}
     OccNet \\ \textbf{Ours}\\
\end{tabular}&
\begin{tabular}{c}
     0.0632\\\textbf{ 0.0087}\\
\end{tabular}&
\begin{tabular}{c}
      0.0363\\\textbf{0.0077}\\
\end{tabular}\\ \hline
THUman2.0 & 
\begin{tabular}{c}
     OccNet \\ \textbf{Ours}\\
\end{tabular}&
\begin{tabular}{c}
     0.0609\\ \textbf{0.0079}\\
\end{tabular}&
\begin{tabular}{c}
     0.0312 \\\textbf{0.0072}\\
\end{tabular} \\ \hline
\end{tabular}
\caption{Quantitative comparison with OccNet~\cite{mescheder2019occupancy}.}
\label{tab:quant_compare_occnet}
\end{table}
\subsection{Qualitative Evaluation}
%
\begin{figure*}
    \centering
  \includegraphics[width=0.9\textwidth]{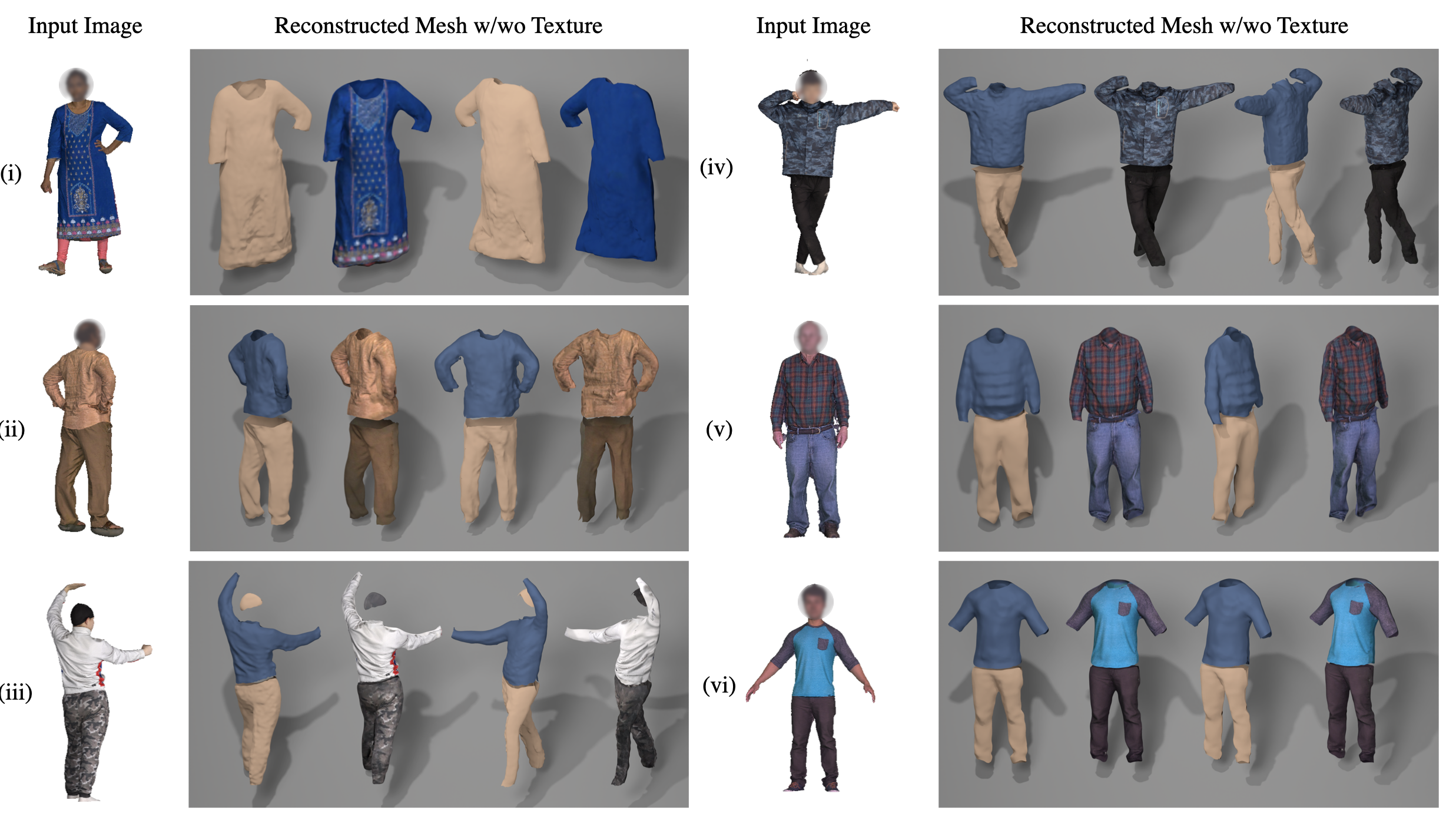}
  \caption{Qualitative results of xCloth on test samples from different datasets. }
  \label{fig:ourresults_on_datasets}
\end{figure*}
We show the qualitative results of xCloth on the test samples taken from three datasets in the \autoref{fig:ourresults_on_datasets}.
As we can infer, we are able to model arbitrary style loose clothing as shown in \autoref{fig:ourresults_on_datasets}$(i)$ and handle self-occlusions as shown in \autoref{fig:ourresults_on_datasets}$(i)$. \autoref{fig:ourresults_on_datasets}$(iii)$ and $(iv)$ demonstrate the ability of our framework to deal with arbitrary poses. 
Additionally, we show the ability of xCloth to reconstruct 3D garments from in-the-wild/catalogue images taken from the internet with arbitrary loose clothing styles in \autoref{fig:ourresults_internet}. 
We also compare our framework with the existing state of the art methods (BCNet \cite{Jiang2020BCNetLB}, MGN \cite{bhatnagar2019mgn}, Pix2Surf \cite{mir20pix2surf} , OccNet \cite{mescheder2019occupancy}) on in-the-wild images taken from the internet. Compared to  the other methods, our framework is superior in recovering high quality textured meshes, as can be seen in \autoref{fig:main_comparison}. Pix2Surf \cite{mir20pix2surf} can transfer texture map for specific cloth categories from two input images (front and back), but it fails to generalize if the input cloth image is not in canonical form. Similarly, \cite{bhatnagar2019mgn} only recovers the geometry of specific categories (Shirt, Tshirt, Shorts, Pants \& Long Coat) as the final output but the texture stitching pipeline of \cite{alldieck19cvpr} is used for transferring texture which requires multi-view images for a decent texture recovery, so we only qualitatively compare only the geometry under monocular settings. We train OccNet \cite{mescheder2019occupancy} on the 3DHumans \cite{3DHumansDataset} and THUman2.0 \cite{yu2021function4d} datasets and show qualitative comparison in \autoref{fig:comparison_occnet}. 
Here ''Not Suppported'' is listed when respective garment template is unavailable.  

It is important to note that, unlike other methods, our framework can recover both fine-grained geometrical details as well as textured representation. Further, our framework can generalize to loose and diverse garments with a complex style, unlike other template-based approaches.
%
\section{Discussion}
\subsection{Ablation Study}
\textbf{Effect of Normal Loss:}
Here, we discuss the importance of adding a decoder branch for predicting normal peelmaps. 
The depth estimation helps recover the global structure of the surface. However, it might not recover high-frequency geometrical details (like wrinkles) on the surface. The surface normals aptly capture such details, and hence predicting normal peel maps (while minimizing average loss) helps in recovering fine geometrical details of the garments.
Additionally, it also helps in regularising the depth map prediction, and yielding lowers values for P2S and NRE. Thus, when we exclude the Normal Loss ($L_{norm}$) and associated loss, the surface quality deteriorates, leading to a drop in P2S and NRE metrics, as reported in \autoref{tab:ablate_normal_loss}.\\
~\\
\textbf{Choice of Architecture:}
Instead of directly adopting the ResNet architecture proposed in \cite{sharp}, we perform an ablative study on the architecture choice. We experiment with U-Net \cite{Ronneberger2015UNetCN} \& Stacked-Hourglass Network \cite{newell2016stacked} apart from ResNet \cite{He2016DeepRL} and compute the P2S error on the final reconstruction while training on the 3DHumans dataset. The corresponding P2S error values for UNet, ResNet and Stacked-Hourglass Network are 0.0062, 0.0074 and \textbf{0.0057}, respectively. Thus, we retain ResNet as the underlying architecture and build on top of it.\\
Please refer to the supplementary material for further ablative experiments. 
%
%
%
\subsection{Analysis \& Insights}
We adopted the non-parametric peeled representation to model garments, thereby removing the need for predefined garment templates which can model only a fixed number of garment styles (mostly tight clothing). The peeled representation is inherently sparse and also deals with self-occlusion. Additionally, the peelmaps (depth, segmentation, normal \& RGB) are pixel-aligned and hence offer an easy way to extract each garment separately from the reconstructed clothed body. SMPL peeled depth prior provided to the common encoder deals with both pose and depth ambiguity in the monocular setup. The proposed novel SMPL body segmentation prior helps the segmentation decoder to localise the body parts while predicting garment segmentation labels (e.g. beanie extracted in \autoref{fig:ourresults_on_datasets}$(iii)$). 
However, it is important to note that our proposed non-parametric framework predicts garment specific segmentation peelmaps for significantly loose clothing styles that are not in direct correspondence to SMPL body segmentation prior.  
Apart from recovering the geometry, peeled representation also guides automatic seams estimation, enabling UV parametrization of any arbitrary garment mesh. 
Another useful insight comes from employing a common encoder with multiple decoder branches. All the different decoder branches provide each other just the relevant information by propagating the gradients via the common encoder while avoiding interfering  directly with each other, resulting in more flexible learning. This is evident from the ablation study of normal loss, where the addition of $D_{norm}$ branch helps to reduce the P2S distance values in the predictions of $D_{depth}$.

For inpainting of the texture atlas, we explored existing learning-based SOTAs. However, almost all inpainting solutions employ generative models, which tend to produce blurry output. Since garment textures come in all kinds of patterns, colors and styles, the unavailability of such diverse datasets to enable generalized learning is another key reason for the failure of learning-based methods. Here, structure completion in inpainting performs effortlessly better than learning-based methods as the former takes cues coming from the structures and repeated patterns present in the image and tries to replicate them periodically.
Please refer to supplementary material for the comparative study.
%
%
\subsection{Limitations \& Future Work}
Our framework does not enforce any spatial relationship between similar parts (e.g. sleeves) of the different garment styles (e.g.shirt and long coat) in the UV space. One possible option is to explore continuous surface embeddings (e.g., CSE~\cite{Neverova2020ContinuousSE}) to establish some meaningful relationship across an independent reconstruction of different garment categories. 
Additionally, in the case of monocular video, our framework is not guaranteed to extract consistent texture for each frame. In future, it would be interesting to explore solutions to predict consistent texture across all the frames. 
Our framework can be further extended to predict the underlying body with accurate shape and pose (instead of taking from SMPL prior) and also to deal with layered clothing, which is extremely difficult to model due to the absence of such real-world datasets.
\section{Conclusion}
We propose a novel method for template-free digitization of 3D garments with automated texture-mapping. We adopt a non-parametric representation and exploit it in a novel way to recover textured 3D garment meshes that can be rendered with a high-fidelity appearance. We compare our framework both qualitatively and quantitatively, where we outperform existing SOTA methods and demonstrate the generalization of our framework to in-the-wild images. We provide an ablative study and also discuss insights gathered from the analysis. Finally, we discuss the limitations \& future directions of our proposed framework.\\
\\
\noindent
\textbf{Acknowledgement: } We thank \href{https://dhawal1939.github.io}{Dhawal Sirikonda} for helping us in the GPU based implementation of texture filling module. 


\clearpage
\vfill\eject
\bibliographystyle{ACM-Reference-Format}
\balance
\bibliography{sample-base}

\end{document}